\documentclass[letterpaper]{article}
\usepackage{aaai24}
\usepackage{times,helvet,courier}
\usepackage[hyphens]{url}
\usepackage{graphicx}
\usepackage{amsfonts}
\usepackage{natbib,caption,booktabs}

\usepackage{amsmath}

\usepackage{algorithm,algorithmic}
\usepackage{newfloat,listings}

\DeclareMathOperator*{\argmin}{arg\,min}

\usepackage{subfigure}

\usepackage{color}
\usepackage[none]{hyphenat}
\usepackage{xcolor}

\DeclareCaptionStyle{ruled}{labelfont=normalfont,labelsep=colon,strut=off} 
\floatstyle{ruled}
\newfloat{listing}{tb}{lst}{}
\floatname{listing}{Listing}

\graphicspath{{./figs/}}

\title{Accelerating Cutting-Plane Algorithms via Reinforcement Learning Surrogates}

\author{
    Kyle~Mana$^{1}$,
    Fernando~Acero$^{1,2}$,
    Stephen~Mak$^{3}$,
    Parisa~Zehtabi$^{1}$,
    Michael~Cashmore$^{1}$,
    Daniele~Magazzeni$^{1}$,
    Manuela~Veloso$^{1}$
}
\affiliations{
    \textsuperscript{\rm 1}J.P. Morgan AI Research\\
    \textsuperscript{\rm 2}University College London\\
    \textsuperscript{\rm 3}University of Cambridge\\
    \{kyle.mana, parisa.zehtabi, michael.cashmore, daniele.magazzeni, manuela.veloso\}@jpmorgan.com, fernando.acero@ucl.ac.uk, sm2410@cam.ac.uk
}

\begin{document}

\maketitle

\begin{abstract}

Discrete optimization belongs to the set of $\mathcal{NP}$-hard problems, spanning fields such as mixed-integer programming and combinatorial optimization. A current standard approach to solving convex discrete optimization problems is the use of cutting-plane algorithms, which reach optimal solutions by iteratively adding inequalities known as \textit{cuts} to refine a feasible set. Despite the existence of a number of general-purpose cut-generating algorithms, large-scale discrete optimization problems continue to suffer from intractability. In this work, we propose a method for accelerating cutting-plane algorithms via reinforcement learning. Our approach uses learned policies as surrogates for $\mathcal{NP}$-hard elements of the cut generating procedure in a way that (i) accelerates convergence, and (ii) retains guarantees of optimality. We apply our method on two types of problems where cutting-plane algorithms are commonly used: stochastic optimization, and mixed-integer quadratic programming. We observe the benefits of our method when applied to Benders decomposition (stochastic optimization) and iterative loss approximation (quadratic programming), achieving up to $45\%$ faster average convergence when compared to modern alternative algorithms. 

\end{abstract}

\section{Introduction}
A large number of problems require discrete decisions. Examples include the decision to purchase an item in whole units, schedule tasks with finite resources, or plan the shortest route through given locations. Even seemingly simple problems can become incredibly challenging to solve when they necessitate discrete decisions \cite{parker2014discrete}. In some cases, discrete optimization problems are provably unsolvable in polynomial time, e.g., it is known that integer programs with quadratic constraints are not solvable at all by Turing machines \cite{jeroslow1973there}. Indeed, problems that would otherwise take fractions of a second to solve can take hours, if not days in discrete space. 

When faced with discrete decisions, heuristic methods such as rounding can offer fast solutions, but at an unknown cost of sub-optimality and if not careful, infeasibility. Due to these issues, there is a strong desire to generate naturally integer, and provably optimal solutions. General-purpose solvers typically rely on a mixture of branch-and-bound and cutting-plane methods to achieve this \citep{bonami2008algorithmic}. Despite advancements in cutting and branching techniques, large-scale problems remain heavily dependent on domain-specific algorithms that exploit problem structure for more efficient convergence. In this paper, we propose a procedure that leverages reinforcement learning for accelerating cutting-plane algorithms. We focus on the domain-specific cutting-plane algorithms of Benders decomposition (applied to stochastic optimization) and a cutting-plane method for solving regression problems with $L_0$ regularization. Our proposed method is generalizable to any cutting-plane procedure and not isolated to the two examples given.

The first of the two cutting-plane procedures we consider, Benders decomposition (BD), is a method that aims to exploit a unique \textit{block} structure commonly found in stochastic optimization (SO) problems. Considering each scenario as a unique block, the global problem is decomposed into a master problem (MP) and a collection of scenario-specific sub-problems (SP). Each SP ingests decisions from the MP, and shares loss information back to the MP in the form of constraints. Iteratively, the MP gains a better understanding of global loss exhibited by each SP, and eventually converges \cite{benders1962partitioning}.

The second procedure we enhance is aimed at solving machine learning problems with sparsity enforced via an $L_0$ regularization term. In settings such as medical imaging \cite{daducci2014sparse}, economics \cite{fan2011sparse}, or causal learning \cite{ide2021cardinality}, sparsity via $L_0$ regularization plays a critical role. Furthermore, it has been shown that regularization techniques such as lasso ($L_1$), ridge ($L_2$), and elastic-net ($L_1, L_2$) can improve out-of-sample performance \citep{zou2005regularization}. Typically, sparse regression is implemented via an $L_1$ regularization term (i.e. lasso), eliminating the necessity to optimize over support of the coefficient set. \citet{louizos2018learning} note that using $L_0$ penalties in parametric models is generally intractable due to non-differentiability and the combinatorial nature of cardinality regularization. Despite its complexity, the importance of $L_0$ regularization is advocated for by \citet{bertsimas2016best}, who argue that using $L_1$ regularization to achieve sparsity results in undesirably biased coefficient estimates by disproportionately affecting larger coefficients in the set. To solve least-squares regression problems with $L_0$ regularization, we introduce a cutting-plane algorithm that iteratively optimizes the sum of penalized support, and a proxy variable bound via sub-gradient approximations of the least-squares loss.

Both of the cutting-plane procedures we consider suffer from three easily observed limitations. First, in discrete space they rely on an $\mathcal{NP}$-hard mixed-integer master problem (MIMP). Second, they operate by abstracting the loss function to a proxy variable, bound from below (assuming minimization without loss of generality) by SP constraints. This can lead to solutions being highly sub-optimal until loss is well represented via constraints. Third, with each iteration an SP generates sub-gradients that are passed to the MIMP as constraints (i.e. cuts), a process which linearly scales the complexity of the MIMP. 

\subsection{Related Work} Acknowledging the overall benefit and potential of cutting-plane algorithms, accelerating these methods has become a compelling research problem. \citet{magnanti1981accelerating} proposed pareto-optimal cut selection for BD. In production routing applications, Adulyasak et al.~\shortcite{BendersProductionRouting} implement lower-bound lifting inequalities to tighten initial lower bounds, and exploit scenario grouping to reduce complexity at each iteration. Crainic et al.~\shortcite{PartialBD} aid initial iterations by including an informative subset of scenarios within the MIMP. Lee et al.~\shortcite{MLBenders} offer a machine learning approach to predict constraint importance; retaining only important cuts and limiting MIMP complexity. Each of these proposals has shown computational benefits, but remain solely dependent on the expensive MIMP to generate successive solutions. In contrast, \citet{HeuristicMaster} replace the MIMP of BD with a genetic algorithm to produce faster feasible solutions. Although the heuristic produces fast MP solutions, it is still reliant on SP approximations to obtain scenario loss, and offers feasible as opposed to certifiably optimal solutions. We refer to \citet{rahmaniani2017benders} for a review of BD methods.

Machine learning methods have been explored for mixed-integer programming and combinatorial optimization (CO). We refer to \citet{mazyavkina2021reinforcement} for a review of RL for CO. \citet{nair2020solving} use neural networks trained via imitation learning to improve branch-and-bound methods for solving MIPs. RL has been used to solve large combinatorial problems, achieving performance close to expert implementations, as shown by Delarue et al.~\shortcite{RLTSP} for notoriously challenging capacitated vehicle routing problems. Recent work has explored the use of RL to improve the performance of modern CO solvers, which typically rely on human-designed heuristics tuned with experience or data. In this sense, RL for cut selection in Integer Programs (IPs) was proposed by Tang et al.~\shortcite{tang2020reinforcement}, and hierarchical RL for cut selection in MILPs was proposed by \citet{wang2023learning}.

 In our work, a surrogate to the MIMP generates fast solutions after learning pseudo-optimal decisions via RL in similar environments. At varying rates, the MIMP is still run to retrieve the certificate of optimality offered. Our contributions are as follows:
\begin{itemize}
    \item A generalized method of accelerating cutting-plane algorithms that retrieves optimal solutions while drastically reducing run times.
    \item Three surrogate solution selection methods, including one that uses cuts to inform selection of surrogate MP solutions, offering a further unification of the surrogate MP within the cutting-plane algorithmic framework.
    \item Empirical evaluation of our approach on two different cutting-plane algorithms. We offer explicit formulations, leverage the learned policy of an RL agent as our surrogate MP in both cases, and provide results showing up to a $45\%$ reduction in run-time against modern alternative methods.
\end{itemize}

\section{Background}
We now discuss relevant background on BD, cutting-planes for $L_0$ regression, and RL.

\subsection{Benders Decomposition}
A widely used form of stochastic optimization is Sample Average Approximation (SAA). In essence, SAA aims to approximate loss over the distribution of possible scenarios using Monte-Carlo simulation. In SAA, $R$ scenarios are simulated, with each simulation yielding its own deterministic SP with a loss function $f(x, w, D_r)$, where $x$ is a set of global decisions (universal across all scenarios), $w$ is a cost vector, and $D_r$ is a set of scenario-specific parameters. The total loss is computed as the average across scenarios,
\begin{equation}\label{eq: generalSP}
    \ell(x) = \frac{1}{R} \sum_{\forall r \in R} f(x, w, D_r)
\end{equation}

 To combat scalability issues as the number of simulations grow, decomposition methods are commonly employed to solve SAA. Here we introduce the principles of Benders decomposition. Consider an SAA problem of the form: 
\begin{equation}
    \min_{x, y} c^Tx + \frac{1}{R}\sum_{\forall r \in R}w^Ty_r
\end{equation}
s.t.
\begin{equation}
    Ax = b
\end{equation}
\begin{equation}
    Bx + D_ry_r = g, \qquad \forall r \in R
\end{equation}
\begin{equation}
    x \in \mathbb{Z}, y_r \in \mathbb{Z}^+ , \qquad \forall r \in R
\end{equation}

where $x$ is our set of global decisions, $A$, $b$, and $B$ are parameters that define constraints on $x$, $c$ is the cost of global decisions, $D_r$ are scenario-specific parameters, $y_r$ is a set of decisions made independently within each scenario, $g$ constrains a combination of global and scenario-specific decisions, and $w$ is the cost of each scenario-specific decision. In this formulation, $w^Ty_r$ is equivalent to \eqref{eq: generalSP}. The first step of BD is to separate global decision variables $x$ and scenario specific decision variables $y_r$, yielding a MP:
\begin{equation}\label{eq: BDMast}
    \{\min_{x, \theta} c^Tx + \frac{1}{R}\sum_{\forall r \in R}\theta_r: Ax = b, x \in \mathbb{Z}^+\}
\end{equation}

and a collection of $R$ SPs, where $\forall r \in R$: % we have:

\begin{equation}\label{eq: BDPrimal}
    \{\min_{y_r} w^Ty_r: D_ry_r = g - Bx^*, y_r \in \mathbb{R}^+\}
\end{equation}

The SPs ingest a fixed $x^*$ based on the solution to \eqref{eq: BDMast}, and are solved to obtain optimal SP decisions $y_r$. Note that BD introduces a set of auxiliary variables $\theta_r, \forall r \in R$ to the MP \eqref{eq: BDMast}. This auxiliary variable, frequently called the recourse variable, is responsible for tracking an approximation of the SP loss that has been moved to \eqref{eq: BDPrimal}. Let us assume the SP is always feasible. This is not a necessary assumption, but simplifies the following description of BD.

Integrality on $y_r$ has been relaxed in the SP. This relaxation is necessary for BD, and is only possible when (i) the SP variables were not discrete to begin with, or (ii) the decomposition results in a totally-unimodular SP structure. Taking the dual of the SP yields:

\begin{equation}\label{eq: BDdual}
    \{\max_{q_r} q_r^T(g-Bx^*): q_r^TD_r \leq w \}
\end{equation}

\begin{figure}[H]
    \centering
    \includegraphics[width=0.95\linewidth]{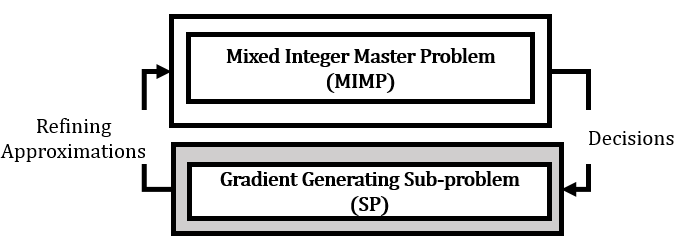}
    \caption{Iterative procedure of Benders decomposition, alternating between a MIMP \eqref{eq: BDMast} and SP \eqref{eq: BDdual}.}
    \label{fig: BDIterative}
\end{figure}

The dual SP has three essential properties. First, through strong duality the optimal value of \eqref{eq: BDdual} is equivalent to the optimal value of \eqref{eq: BDPrimal} at $x^*$. Second, the objective function \eqref{eq: BDdual} is linear with respect to the MP decisions $x$. And lastly, with the optimal dual values of $q^*_r$ we can establish
\begin{multline}
    \{\min_{y_r} w^Ty_r: D_ry_r = g - Bx\} \geq \\ q_r^{*T}(g-Bx), \forall x \in \mathbb{R}, \forall w \in \mathbb{R}
\end{multline}
via weak duality.
With these traits established, we see that the optimal dual SP objective $q_r^{*T}(g-Bx)$ can be included as a valid constraint on $\theta_r$ in the MIMP. These constraints serve as sub-gradient approximations of the SP loss. For each SP solution, we can update the MIMP with the valid constraint of $\theta_r \geq q_r^{*T}(g-Bx)$ and re-solve for a new $x$. This process is repeated until the SP's do not offer any strengthening constraints on $\theta_r$, indicating convergence and full approximation of SP loss. Figure \ref{fig: BDIterative} offers a visual representation of this process, which translates to the $L_0$ regularization application that we introduce next.

\subsection{Cutting-Planes for $\boldsymbol{L_0}$ Regularized Regression}

In statistical analysis and machine learning, high-dimension datasets may necessitate the use of methods that distinguish a meaningful feature subset. Sparse regression aims to minimize loss while limiting support over the coefficient set, represented by $\beta$. Using $L_0$ regularization encourages this sparsity by penalizing support over $\beta$. Consider data given by the general form $y = f(X,\beta) + \epsilon$, where $f$ is a possibly nonlinear function. The problem of $L_0$ regularized regression

\begin{equation}
\label{eq: L0regularizedregression}
    \min_{\beta} ||f(X,\beta) - y||_2^2 + \lambda \sum_{\forall i \in p} ||\beta_i||_0
\end{equation}

can be cast as a mixed-integer quadratic program MIQP, where $P$ represents the number of features and $M$ is a sufficiently large constant:

\begin{multline}
\label{eq: regressionMIP}
    \{\min_{\beta, z} ||f(X,\beta) - y||_2^2 + \lambda \sum_{\forall i \in p} z_i: |\beta_i| \leq z_i M\text{ } \forall i \in P, \\
    \beta \in \mathbb{R}^P, z \in \{0,1\}^P \}
\end{multline}

MIQPs of the form \eqref{eq: regressionMIP} struggle in high-dimensional settings, inspiring our use of cutting-plane procedures. To do this, we can reframe \eqref{eq: regressionMIP} as a linear program:

\begin{multline}\label{eq: RegressionMP}
    \{\min_{\beta, \theta, z} \theta + \lambda \sum_{\forall i \in p} z_i: |\beta_i| \leq z_i M \text{ } \forall i \in P, \theta \geq 0, \\
    \beta \in \mathbb{R}^P, z \in \{0,1\}^P, \theta \in \mathbb{R} \}
\end{multline}

where $\theta$ serves as a proxy variable for the convex and differentiable loss $||f(X,\beta) - y||_2^2$. For each iteration $n$ of \eqref{eq: RegressionMP} we compute the loss $l_n$ and sub-gradient $\nabla g(\beta^{(n)})$ to constrain $\theta$ with a lower bound in the form of a linear constraint. If we denote the polytope defined by \eqref{eq: RegressionMP} as $\mathcal{P}_0$, after $n$ iterations the polytope $\mathcal{P}_n$ is restricted to $\mathcal{P}_{n-1} \cup \{\beta, \theta: \theta \geq l_n + \nabla g(\beta^{(n)}) (\beta - \beta^{(n)})\}$. This process terminates when the gap between $l_n$ and the evaluation of the objective from \eqref{eq: RegressionMP} over $\mathcal{P}_n$ is within a tolerance $e$.

\subsection{Reinforcement Learning} 
RL is a framework for solving sequential decision-making problems, formulated as a Markov Decision Process (MDP) \cite{sutton_reinforcement_2018}. Our proposed framework requires casting the optimization problem at hand as an MDP, i.e., decision variables form the action space and the negative of the cost function is the reward function. Formally, MDPs are defined as a 4-tuple $\langle \mathcal{S}, \mathcal{A}, \mathcal{T}, \mathcal{R} \rangle$ where $\mathcal{S}$ is the state space, $\mathcal{A}$ is the action space, $\mathcal{T}$ is the set of transition probabilities from states $s_t$ to $s_{t+1}$ upon taking action $a_t$, and $\mathcal{R}$ is the reward function. We note how $\mathcal{T}$ and $\mathcal{R}$ may be non-deterministic, and therefore MDPs may be used to model problems pertaining SO. A discount factor $\gamma$ is typically introduced to discount rewards. We denote a policy parametrised by $\phi$ as $\pi_{\phi}:\mathcal{S} \rightarrow \mathcal{A}$.

RL algorithms may be categorized as value-based or policy gradient methods. Value-based methods learn a value function or action-value function, from which optimal actions can be implicitly obtained, whereas policy gradient methods directly optimize an explicit representation of the optimal policy. The value function and action-value functions for episodic MDPs with horizon $T$ are given by:
\begin{equation}
    V^{\pi}(s) = \mathbb{E}_{\pi}\left[\sum_{k=t}^{T} \gamma^k r_{t+k+1}|s=s_t\right]
    %V^{\pi}(s, a) = \max_{a \in \mathcal{A}} Q^{\pi}(s, a)
\end{equation}
\begin{equation}
    Q^{\pi}(s, a) = \mathbb{E}_{\pi}\left[\sum_{k=t}^{T} \gamma^k r_{t+k+1}|s=s_t, a=a_t\right]
\end{equation}

where it can be seen that the action-value function $Q^{\pi}$ is only practically applicable for discrete action spaces (or discretizations of continuous action spaces), and is directly susceptible to the curse of dimensionality. Alternatively, policy gradient methods update policy parameters $\phi$ via an estimate of the policy gradient, as first introduced by \citet{SuttonPolicy}. Policy gradient methods can be used for discrete or continuous action spaces, and are based on some expression of the policy gradient:
\begin{equation}
    \nabla_{\phi} \mathbb{E} \left[\sum_{t=0}^{T} r_t \right] \approx \mathbb{E} \left[ \sum_{t=0}^{T} \Psi_t \nabla_{\phi} \log \pi_{\phi}(a_t | s_t) \right]
    \label{eq:policygradient}
\end{equation}
where $\Psi_t$ may be the (discounted) returns of the trajectory, the action-value function, the advantage function, temporal-difference residual, or else, yielding different policy gradient algorithms \cite{Schulmanetal_ICLR2016}. Proximal Policy Optimization (PPO) is a powerful policy gradient algorithm that avoids detrimentally large policy updates proposed by \citet{PPOSchulman}, where a surrogate objective to \eqref{eq:policygradient} is used based on a probability ratio which is clipped whenever $|\frac{\pi_{\phi}^{\text{new}}(a_t | s_t)}{\pi_{\phi}^{\text{old}}(a_t | s_t)}| > \epsilon$ for some small $\epsilon$, providing a lower bound on the unclipped objective. Consequently, PPO provides stable updates, whilst being on-policy and thus susceptible to low sample efficiency compared to off-policy algorithms such as Q-learning. Compared to value-based methods, using an explicit parametric policy is a more natural choice for solving MDPs for which the optimal policy may be stochastic, as the policy can be a parametric stochastic function, whereas value-based methods require crafting a sampling strategy (e.g. $\epsilon$-greedy) to generate a stochastic policy from a value function. \citet{SuttonPolicy} discussed advantages of stochastic policies in contrast to policies induced by value functions. % when introducing policy gradient methods.

Actor-critic methods combine the benefits of value-based methods and policy gradients. Value estimates can be used as a baseline for advantage estimates \cite{SuttonPolicy}. Modern actor-critic algorithms frequently use Generalized Advantage Estimation (GAE), an exponentially-weighted advantage estimator that addresses the bias-variance tradeoff \cite{Schulmanetal_ICLR2016}. We use actor-critic PPO with GAE.

\section{Accelerating Cutting-Plane Algorithms}
We now introduce our proposed acceleration method. First, we offer specifics on how a surrogate is used in place of the MIMP. Then, we introduce three possible mechanisms for leveraging the surrogate. Lastly, we offer a more thorough coverage of the theoretical benefits that may be provided by a surrogate, and known deficiencies of cutting-plane methods that it addresses. We focus on surrogates learned via RL, but we note that our proposed method is agnostic to the nature of the method used to learn the surrogate policy. 

\subsection{Surrogate-MP}

\begin{figure}[t]
    \centering
    \includegraphics[width=0.95\linewidth]{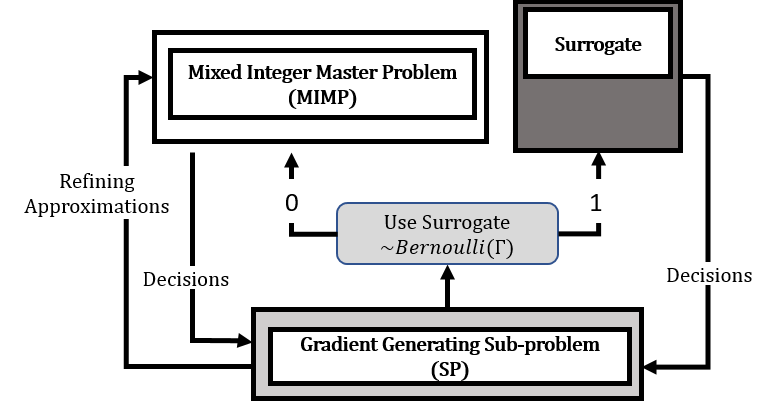}
    \caption{Iterative procedure of Surrogate-MP.}
    \label{fig: BDRLIterative}
\end{figure}

Recall the iterative procedure outlined in Figure \ref{fig: BDIterative}. As is the case in our two examples, we assume the sub-gradient can be computed efficiently (as a linear problem in the case of BD, and in closed form in the case of $L_0$ regularization). However, each case calls back to an $\mathcal{NP}$-hard MIMP, with complexity that scales linearly with the number of iterations. Given these dynamics, there is a strong desire to (i) increase the speed of each MP iteration and (ii) decrease the total number of calls to the MIMP required. We achieve both results by periodically introducing a faster surrogate in place of the MIMP (Figure \ref{fig: BDRLIterative}). This surrogate can be any policy that has learned to map the input space to the discrete decision space with the objective of minimizing the problem loss or cost. 
 
Note in this modified schema that with each iteration, the decision to use the surrogate in place of the MIMP is drawn from a Bernoulli distribution with a control parameter $\Gamma$. If a value of 1 is returned from the Bernoulli distribution, the surrogate is used to generate global decisions. Otherwise, the standard MIMP is run and the optimality gap can be confirmed. Regardless of whether the MIMP or surrogate are used, global decisions are passed to the SP and loss approximating cuts are added to the iterative process.

\subsection{Leveraging Surrogate Solutions}
The solutions produced by a surrogate can be used in a variety of ways, and we propose three selection mechanisms. These variants are aimed at answering: (i) How can we use the surrogate to improve convergence? (ii) If surrogate actions are non-deterministic, how can we decide which actions are best to use? The three methods we propose are greedy selection, weighted selection, and informed selection. Each of these methods assume the surrogate has generated a batch of stochastic trajectories of actions for $B$ episodes, i.e. a batch of $B$ distinct solutions, each with loss $\ell_b$ equal to the negative returns of the trajectory in the MDP. 

\paragraph{Greedy Selection} This method selects the best performing solution within a batch, i.e. $\argmin_b(\ell_b)$. In the case of BD we evaluate the solution against an expected outcome, as performance cannot be deterministically evaluated.

\paragraph{Weighted Selection} Rather than selecting actions based on expected performance, we can perform weighted random sampling. We use the loss of each solution $\ell_b$ to define a probability mass $p(b) = \frac{\frac{1}{\ell_b}}{\sum_{\forall b \in B}\frac{1}{\ell_b}}$ for random sampling.

\paragraph{Informed Selection} The final proposal incorporates feedback from the constraint matrix on $\theta$ at a given iteration. The benefit of utilizing the constraint matrix to select surrogate solutions is that these constraints inherently motivate exploration to either (i) minimal or (ii) poorly approximated regions of the convex loss. Given final convergence is defined by a binding subset of these constraints, it is desirable to explore these regions.

We introduce the constraint matrix $A_r \in \mathbb{R}^{I \times N}$ which contains the sub-gradient approximations imposed on $\theta_r$, and a row vector of constant values $c_r \in \mathbb{R}^{B}$ that is added to each sub-gradient approximation. Recall $r \in R$, and $R$ is the number of scenarios for BD, while $R=\{1\}$ for $L_0$ regularization. $I$ refers to the iteration number of the cut-generating algorithm, and $N$ refers to the number of MP decision variables. Each iteration generates a new set of sub-gradient approximations which are added to $A_r$. These are the same sub-gradients that are applied to $\theta_r$ in the MP, and are generated using the SP. On a given iteration, we have a batch of $B$ solutions that have been generated by the surrogate. Decisions for this batch are represented by matrix $D \in \mathbb{Z}^{N \times B}$. We begin by computing the loss approximations of each gradient, for each of the $B$ solutions. This is given by $\mathcal{L}_{r} \in \mathbb{R}^{I \times B}$, which we define as:
\begin{equation}
    \mathcal{L}_{r} = A_r \cdot D + (\mathbf{c_r} \cdot 1^{1 \times I})^T 
\end{equation}

The $\mathcal{L}_{r}$ matrix contains approximations of the SP loss for each of the $B$ solutions, generated by each of the $I$ constraints currently placed on $\theta_r$. We can now take the maximum value for each column $B$ as the approximated cost of solution $b$. In LP terms, this maximum value relates to the binding constraint on $\theta_r$ in the MIMP, and is thus our best approximation of SP cost at that point. We represent this approximation ($\hat{\ell}_{b,r}$) as: 

\begin{equation}
    \hat{\ell}_{b,r} = \max_{\forall i \in I} (\mathcal{L}_{r})_{i,b}
\end{equation}

Now we fully approximate the expected loss for each of the $M$ solutions by taking an average across all $r \in R$, and adding the fixed loss of that decision (denoted $f_b$):

\begin{equation}
    \hat{\ell_b} = \frac{1}{R} \sum_{\forall r \in R} \hat{\ell}_{b,r} + f_b
\end{equation}

The informed selection then solves the problem $\argmin_b \hat{\ell_b}$, which is taken as our MP solution, and passed to the SP for constraint generation.

\subsection{Benefits of Surrogate-MP}
The benefits of using a learned surrogate in place of the MIMP are based on two key observations: 
\begin{enumerate}
    \item The time required to generate solutions from a learned surrogate (e.g. inference on a neural net) is negligible compared to the time required to solve a large-scale MIP.
    \item The surrogate has learned to produce actions from past experience. As a result, SP loss is expressed in surrogate solutions regardless of how well $\theta_r$ approximates SP loss. This means that even at early iterations, the surrogate solutions will be highly reflective of SP loss.
\end{enumerate}

The first benefit is fairly self-explanatory; we desire faster MP solutions, and the surrogate provides them. The second benefit  is more nuanced and worth expanding. We recall the general form MIMP \eqref{eq: BDMast}, where $\theta_r$ offers an approximation of SP loss that is refined through linear constraints generated by \eqref{eq: BDdual}. It is well observed that this approximation can converge quickly if global decisions are localized to the optimal region, but it can also be very slow if global decisions are far from the optimal region or if cuts poorly approximate the loss \cite{PartialBD, StableLocalBenders}. At initialization, $\theta_r$ has not received any feedback from the SP, and is instead bound by some heuristic or known lower bound (commonly $\theta_r \geq 0$ for non-negative loss). Given the lack of information initially imparted on $\theta_r$, the MP generates global solutions that lack consideration of SP loss and can be very distant from the optimal region. Similar to a gradient based algorithm with a miss-specified learning rate, this can lead cutting-plane methods such as BD to oscillate around the minimal region or converge slowly, wasting compute and adding complexity with minimal benefit to the final solution \cite{StableLocalBenders}.

The surrogate mitigates this major issue by generating global decisions that reflect an understanding of their associated SP loss without requiring the MP to have strong loss approximations on $\theta_r$. As a result, initial global decisions generated by the surrogate are localized to the minimal region and cuts can quickly approximate the minimum of the convex loss. These two fundamental benefits are the basis for a $30\%$-$45\%$ reduction in run-times, observed in experiments within the two domains that follow. 

\section{Experiments}\label{experiments} 

We evaluate our proposed acceleration method on two distinct cutting-plane algorithms in separate domains. The first application displays an acceleration of Benders decomposition, using a stochastic inventory management problem consisting of a basic two-stage decision process, reflecting the MP/SP structure displayed in Figure \ref{fig: BDIterative}. The second algorithm is an $L_0$ regularized regression problem as described in the Background section.

\textbf{Inventory Management Problem} (IMP). In the proposed IMP, we assume the required solutions must (i) choose a delivery schedule from a finite set, (ii) decide an order-up-to amount (where order equals order-up-to minus current inventory) for each scheduled day, and (iii) place costly emergency orders if demand cannot be met with current inventory at any time. We assume there is a requirement to satisfy all demand using either \textit{planned schedules}, or more costly \textit{just-in-time emergency orders}. Demand estimates are generated using a forecast model with an error term from an unknown probability distribution.

To model the IMP as a SO mixed-integer problem we introduce the following notation: let $T$ be the set of days $t$, $R$ the set of scenarios $r$,  $S$ the set of schedules $s$, holding cost of an item (per unit-of-measure, per day) $h$, cost of emergency services (per unit) $e$, penalty applied to over-stocking (per unit over-stocked) $q$, and fixed cost of a schedule $f_s$ . The decision space is defined by seven sets of variables, some of which are: the units of holding space required to stock the inventory $p_{tr} \in \mathbb{Z}^+$, the required emergency order quantity $o_{tr} \in \mathbb{Z}^+$, and the number of units that inventory is over-filled by $v_{tr} \in \mathbb{Z}^+$ (all defined $\forall t \in T, \forall r \in R$). The general formulation of our IMP is:

\begin{equation}\label{eq: mipobjectiveIMP}
    \min \sum_{\forall s \in S}(u_s f_s) + \frac{1}{R} \sum_{\forall r \in R} \sum_{\forall t \in T}(p_{tr}h + o_{tr}e + v_{tr}q)
\end{equation}

subject to several constraints. Of primary importance is a set of constraints that link the volume of inventory on hand (in each SP) to the scheduling and order-up-to decisions (MP). The resultant decomposition consists of a master problem objective 
\begin{equation}\label{eq: mipobjectiveMP}
    \min \sum_{\forall s \in S}(u_s f_s) + \sum_{\forall r \in R} \theta_r
\end{equation}
and sub-problem objective 
\begin{equation}\label{eq: mipobjectiveSP}
    \min \sum_{\forall t \in T}(p_{tr}h + o_{tr}e + v_{tr}q),  \forall r \in R
\end{equation}
The SP decisions are constrained by MP scheduling and order-up-to decisions, and we take its dual to generate cuts on $\theta_r$ in the MP. See Appendix for full details. 

As previously mentioned, we leverage policies learned via RL as our Surrogate MPs for both problems. In order to do so, we cast the problems into MDP formulations, and train neural networks (multi-layer perceptrons) as our policy and value functions. The policy networks return log-odds for discrete actions from which stochastic actions are sampled. We now discuss the formulation for the IMP problem, see Appendix for further details. The MDP transitions by selecting a schedule first, and subsequently setting order quantities as required by the schedule, reaching termination at the end of the temporal horizon. The state space includes balance, capacity, current day, cost parameters, forecasted demand (mean and standard deviation) per day, residuals (between forecasted and actual demand) per day, schedule selected, previous order quantities, and which day the agent is ordering for. The action space includes the schedule decision, and the order quantity decision. Note that there is a hierarchical structure to our IMP problem, in that a schedule decision must be made at the beginning of the horizon, and subsequently only order quantity decisions are to be made for each day within the horizon. The reward function is given by the negative of the cost defined in \eqref{eq: mipobjectiveIMP}. We use action masking on the policy network outputs in order to enforce constraints on scheduling and order quantity decisions as required during MDP transitions.

We perform experiments on 153 independent cases of our IMP using real-world data. Each experiment was performed with the following parameters: 500 scenarios ($R=500$), 28 day horizon ($T=28$), and 169 possible schedules ($S=169$). The baseline implementation of BD includes accelerations such as scenario group cuts (Adulyasak et al. \shortcite{BendersProductionRouting}) and partial decomposition (Crainic et al. \shortcite{PartialBD}). We do not compare against a generic implementation of BD due to tractability limitations. 

\textbf{Regularized Regression} (RR). In this problem, the objective is to minimize the sum of a convex loss function and the penalized support of a feature set. We follow the cutting-plane procedure for $L_0$ regularized regression outlined in the Background section, computing $\nabla g(\beta)$ in closed form for the selected sparse feature set. In our experiments, we focus on linear regression problems, i.e. $f(X,\beta) = X^T \beta$ in equation \ref{eq: L0regularizedregression}. %Further details in Appendix.

The MDP is formulated to allow for selecting one feature at a time. The state space includes coefficients $\beta$ for each feature assuming all features were to be used for regression and their corresponding p-values (which do not change as the MDP transitions), as well as coefficients for the currently chosen sparse feature set $\beta\vert_z$ and the corresponding multi-hot encoding $z$ for the currently chosen features (which do change as the MDP transitions). The action space consists of the categorical distribution for all features. The reward function is given by the change in the MSE of the residuals $(||f(X,\beta\vert_z) - y||_2^2)$ at each step as $z$ grows. We use action masking to mask previously chosen features within an episode. The episode terminates when the increase in explained variance when adding a feature is less than the increase in $L_0$ penalty for the added feature.

We perform experiments on 250 regularized regression problems using synthetic data $y = X^T \beta + \epsilon$ with Gaussian noise $\epsilon$ (data generation process is outlined in Appendix).

\section{Results}

We evaluate performance against a baseline for each problem. A modern accelerated version of BD is used as a benchmark for the IMP, and the cutting-plane algorithm we describe in the Background section is used for RR. All MIPs and LPs are solved using the CPLEX commercial solver. Experiments were run on a 36 CPU, 72 GB RAM Linux machine. For every implementation of Surrogate-MP, we deactivate the surrogate after the optimality gap is less than $5\%$, to focus on retrieval of optimality using the MIMP.

\begin{figure}[H]
    \centering
    \subfigure[Inventory Management]{\includegraphics[width=0.49\linewidth]{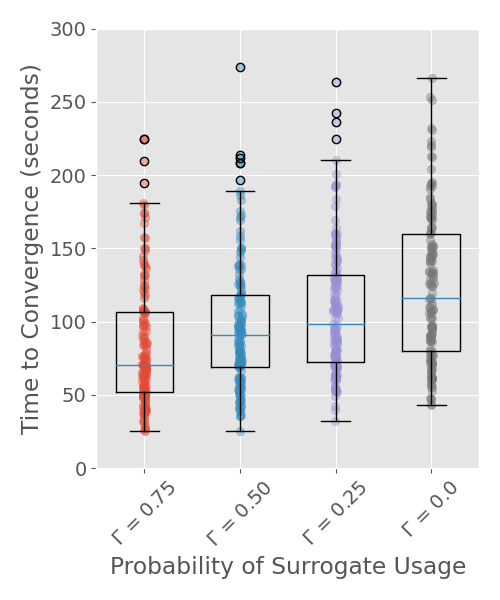}} 
    \subfigure[Regularized Regression]{\includegraphics[width=0.49\linewidth]{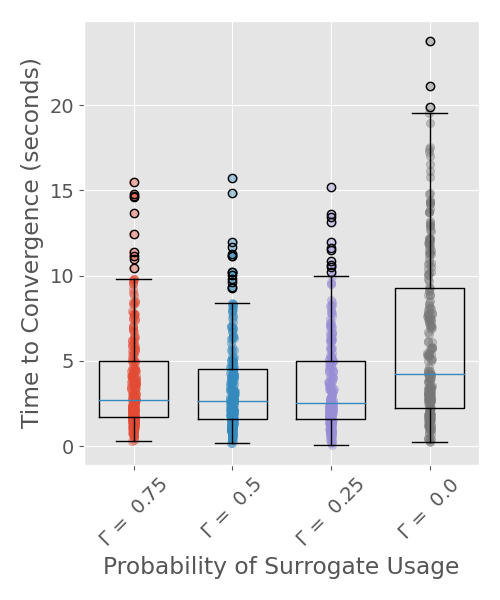}}
    \caption{Convergence instances of BD accelerated by an informed Surrogate-MP, with different surrogate usages.}
    \label{fig: InformedII}
\end{figure}

\begin{figure}[H]
    \centering
    \subfigure[Inventory Management]{\includegraphics[width=0.49\linewidth]{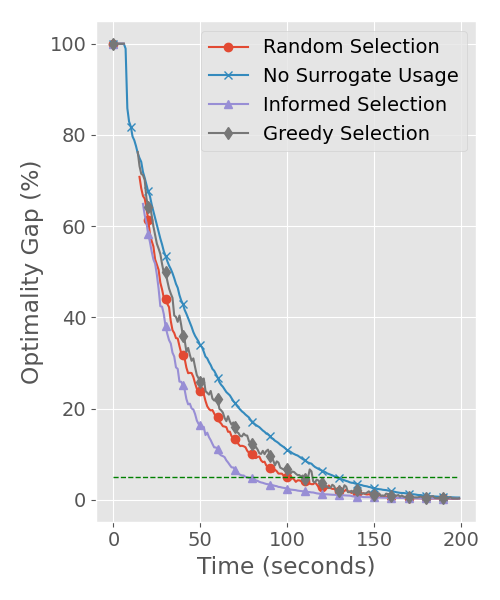}}
    \subfigure[Regularized Regression]{\includegraphics[width=0.49\linewidth]{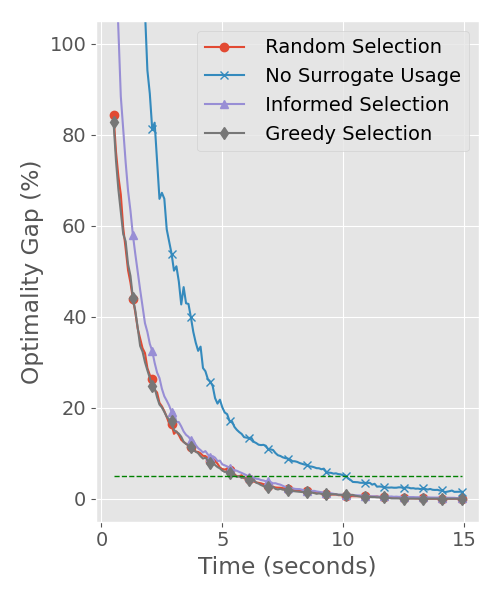}}
    \caption{Convergence rates of a baseline BD, and Surrogate-MP with three selection methods (greedy, weighted, informed). Results generated with $\Gamma=0.75$.}
    \label{fig: CompareI}
\end{figure}

\begin{figure}[H]
    \centering
    \subfigure[Inventory Management]{\includegraphics[width=0.49\linewidth]{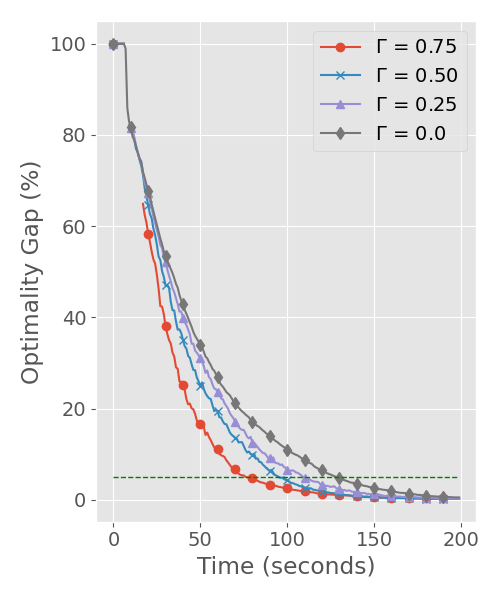}}
    \subfigure[Regularized Regression]{\includegraphics[width=0.49\linewidth]{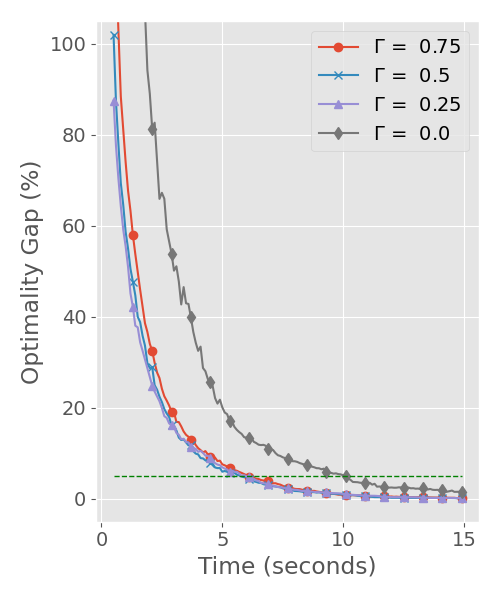}}
    \caption{Convergence rates of BD accelerated by an informed Surrogate-MP with different surrogate usages. Surrogate-MP deactivated at $5\%$ as indicated by dotted line.}
    \label{fig: InformedI}
\end{figure}

\textbf{IMP}. All three selection methods, when ran with $\Gamma=0.75$, produced faster convergence than the baseline model: weighted selection performed $14.96\%$  faster than the baseline, greedy selection achieved $19.43\%$ faster performance, and informed selection performed $30.45\%$ faster. Furthermore, with informed selection Surrogate-MP performed faster on over $88\%$ of instances (convergence rates in Figure \ref{fig: CompareI}, instances of faster convergence in Table \ref{tab:best-counts}). To further investigate the strong performance of informed selection, we experimented with $\Gamma = 0.25, 0.5, \text{and } 0.75$. $\Gamma = 0.75$ was fastest at $30.45\%$ acceleration, $\Gamma = 0.5$ converged $21.24\%$ faster, and $\Gamma = 0.25$ realized $11.84\%$ faster convergence (Figure \ref{fig: InformedII}, Figure \ref{fig: InformedI}). Empirically, RL solutions produced results with 14.73\% higher cost than the optimal results produced by Benders decomposition at convergence (average across all 153 instances). This indicates the RL heuristic solutions are pseudo-optimal, whilst being significantly faster to obtain.

\textbf{RR}.
For $\Gamma=0.75$, greedy selection resulted in $45.31\%$ acceleration, weighted selection in $44.41\%$, and informed selection in $37.97\%$. It is worth noting that RR is a less partially observable MDP than the IMP is. This stronger observability may explain why the RL surrogate benefits less from information sharing via the constraint matrix $A_r$. The convergence rates for each selection method are shown in Figure \ref{fig: CompareI}. For consistency, we vary the $\Gamma$ parameter with informed selection and inspect its impact on convergence. In contrast with the IMP, we observe $\Gamma = 0.5$ results in fastest convergence, accelerating RR by $42.94\%$ (Figure \ref{fig: InformedII}). For RR, varying $\Gamma$ produced much less variance in convergence than in the IMP. Similar to the IMP experiments, we observe Surrogate-MP has outperformed the baseline algorithm across most RR instances: Surrogate-MP achieved better convergence rates on up to $85.60\%$ of instances (Table \ref{tab:best-counts}). Lastly, we compare $L_0$ vs $L_1$ regularization with an $L_1$ penalty ($\lambda$) of 0.1 and 0.5. The two $L_1$ parameter settings are intended to achieve accurate parameter estimation, and effective coefficient recovery (respectively). With $\lambda=0.1$ for both $L_1$ and $L_0$, $L_0$ yields significantly improved coefficient recovery (by 81\%), and parameter estimation (by 34\%), while $L_1$ yields marginally better MSE (by 2\%). With $\lambda=0.5$, $L_1$'s coefficient recovery improves, but both MSE and parameter estimation degrade considerably. These results display the benefits of $L_0$ regularization with respect to unbiased feature reduction. Metric definitions and a table with full results can be found in the Appendix.

\begin{table}[]
\centering
\begin{tabular}{@{}lll@{}}
\toprule
                       & Surrogate-MP   & No Surrogate \\ \midrule
Inventory Management   & \textbf{135 (88.24\%)}  & 18 (11.76\%) \\
Regularized Regression & \textbf{214 (85.60 \%)} & 36 (14.40\%) \\ \bottomrule
\end{tabular}
\caption{Instances of faster convergence when using RL-surrogate with the best performing configuration ($\Gamma$ and selection method), compared to baseline of no surrogate. }
%\vspace{-1mm}
\label{tab:best-counts}
\end{table}

\section{Conclusion \& Future Work}

We use RL to learn surrogates in place of MIMPs for accelerating cutting-plane algorithms, achieving drastic reduction in convergence times. Our proposed method is application agnostic, retrieves certificates of optimality, and can utilize any surrogate capable of generating MP solutions. We provide formulations in two different domains -- Benders decomposition applied to inventory management and a cutting-plane algorithm for $L_0$ regularized regression -- and provide results showing superiority of our approach in $88.24\%$ and $85.60\%$ of instances with a $30.45\%$ and $45.31\%$ reduction in average run-time respectively.

A promising direction for future work would be to design stronger integration between the surrogate, SP, and MP. Our informed selection method is a first step in this direction, and realized promising results. Some additional opportunities we leave unexplored would be to directly inform the surrogate on the strength of past solutions, offer sub-gradient information as a feature, or redesign the surrogate's objective function to reward the strength of subsequent cuts as opposed to mirroring the MP objective directly. We are additionally eager to observe the performance of Surrogate-MP on other cutting-plane algorithms and optimization problems.

\section{Acknowledgments}
We thank Alec Koppel for helpful comments on this manuscript, and anonymous reviewers of this submission. 
\paragraph{Disclaimer.}
This paper was prepared for informational purposes by
the Artificial Intelligence Research group of JPMorgan Chase \& Co. and its affiliates (``JP Morgan''),
and is not a product of the Research Department of JP Morgan.
JP Morgan makes no representation and warranty whatsoever and disclaims all liability,
for the completeness, accuracy or reliability of the information contained herein.
This document is not intended as investment research or investment advice, or a recommendation,
offer or solicitation for the purchase or sale of any security, financial instrument, financial product or service,
or to be used in any way for evaluating the merits of participating in any transaction,
and shall not constitute a solicitation under any jurisdiction or to any person,
if such solicitation under such jurisdiction or to such person would be unlawful.

\bibliography{aaai24}

\begin{thebibliography}{26}
\providecommand{\natexlab}[1]{#1}

\bibitem[{Adulyasak, Cordeau, and Jans(2015)}]{BendersProductionRouting}
Adulyasak, Y.; Cordeau, J.-F.; and Jans, R. 2015.
\newblock Benders Decomposition for Production Routing Under Demand Uncertainty.
\newblock \emph{Operations Research}, 63(4): 851--867.

\bibitem[{Baena, Castro, and Frangioni(2020)}]{StableLocalBenders}
Baena, D.; Castro, J.; and Frangioni, A. 2020.
\newblock Stabilized Benders Methods for Large-Scale Combinatorial Optimization, with Application to Data Privacy.
\newblock \emph{Management Science}, 66(7): 3051--3068.

\bibitem[{Benders(1962)}]{benders1962partitioning}
Benders, J.~F. 1962.
\newblock Partitioning procedures for solving mixed-variables programming problems.
\newblock \emph{Numerische Mathematik}, 4: 238--252.

\bibitem[{Bertsimas, King, and Mazumder(2016)}]{bertsimas2016best}
Bertsimas, D.; King, A.; and Mazumder, R. 2016.
\newblock Best subset selection via a modern optimization lens.

\bibitem[{Bonami et~al.(2008)Bonami, Biegler, Conn, Cornu{\'e}jols, Grossmann, Laird, Lee, Lodi, Margot, Sawaya et~al.}]{bonami2008algorithmic}
Bonami, P.; Biegler, L.~T.; Conn, A.~R.; Cornu{\'e}jols, G.; Grossmann, I.~E.; Laird, C.~D.; Lee, J.; Lodi, A.; Margot, F.; Sawaya, N.; et~al. 2008.
\newblock An algorithmic framework for convex mixed integer nonlinear programs.
\newblock \emph{Discrete optimization}, 5(2): 186--204.

\bibitem[{Crainic et~al.(2016)Crainic, Rei, Hewitt, and Maggioni}]{PartialBD}
Crainic, T.~G.; Rei, W.; Hewitt, M.; and Maggioni, F. 2016.
\newblock \emph{Partial Benders decomposition strategies for two-stage stochastic integer programs}, volume~37.
\newblock CIRRELT.

\bibitem[{Daducci et~al.(2014)Daducci, Van De~Ville, Thiran, and Wiaux}]{daducci2014sparse}
Daducci, A.; Van De~Ville, D.; Thiran, J.-P.; and Wiaux, Y. 2014.
\newblock Sparse regularization for fiber ODF reconstruction: From the suboptimality of $\ell$2 and $\ell$1 priors to $\ell$0.
\newblock \emph{Medical Image Analysis}, 18(6): 820--833.

\bibitem[{Delarue, Anderson, and Tjandraatmadja(2020)}]{RLTSP}
Delarue, A.; Anderson, R.; and Tjandraatmadja, C. 2020.
\newblock Reinforcement learning with combinatorial actions: An application to vehicle routing.
\newblock \emph{Advances in Neural Information Processing Systems}, 33: 609--620.

\bibitem[{Fan, Lv, and Qi(2011)}]{fan2011sparse}
Fan, J.; Lv, J.; and Qi, L. 2011.
\newblock Sparse high-dimensional models in economics.
\newblock \emph{Annu. Rev. Econ.}, 3(1): 291--317.

\bibitem[{Id{\'e} et~al.(2021)Id{\'e}, Kollias, Phan, and Abe}]{ide2021cardinality}
Id{\'e}, T.; Kollias, G.; Phan, D.; and Abe, N. 2021.
\newblock Cardinality-regularized hawkes-granger model.
\newblock \emph{Advances in Neural Information Processing Systems}, 34: 2682--2694.

\bibitem[{Jeroslow(1973)}]{jeroslow1973there}
Jeroslow, R.~C. 1973.
\newblock There cannot be any algorithm for integer programming with quadratic constraints.
\newblock \emph{Operations Research}, 21(1): 221--224.

\bibitem[{Lee et~al.(2021)Lee, Ma, Yu, and Dai}]{MLBenders}
Lee, M.; Ma, N.; Yu, G.; and Dai, H. 2021.
\newblock Accelerating Generalized Benders Decomposition for Wireless Resource Allocation.
\newblock \emph{IEEE Transactions on Wireless Communications}, 20(2): 1233--1247.

\bibitem[{Louizos, Welling, and Kingma(2018)}]{louizos2018learning}
Louizos, C.; Welling, M.; and Kingma, D.~P. 2018.
\newblock Learning Sparse Neural Networks through L\_0 Regularization.
\newblock In \emph{International Conference on Learning Representations}.

\bibitem[{Magnanti and Wong(1981)}]{magnanti1981accelerating}
Magnanti, T.~L.; and Wong, R.~T. 1981.
\newblock Accelerating Benders decomposition: Algorithmic enhancement and model selection criteria.
\newblock \emph{Operations Research}, 29(3): 464--484.

\bibitem[{Mazyavkina et~al.(2021)Mazyavkina, Sviridov, Ivanov, and Burnaev}]{mazyavkina2021reinforcement}
Mazyavkina, N.; Sviridov, S.; Ivanov, S.; and Burnaev, E. 2021.
\newblock Reinforcement learning for combinatorial optimization: A survey.
\newblock \emph{Computers \& Operations Research}, 134: 105400.

\bibitem[{Nair et~al.(2020)Nair, Bartunov, Gimeno, Von~Glehn, Lichocki, Lobov, O'Donoghue, Sonnerat, Tjandraatmadja, Wang et~al.}]{nair2020solving}
Nair, V.; Bartunov, S.; Gimeno, F.; Von~Glehn, I.; Lichocki, P.; Lobov, I.; O'Donoghue, B.; Sonnerat, N.; Tjandraatmadja, C.; Wang, P.; et~al. 2020.
\newblock Solving mixed integer programs using neural networks.
\newblock \emph{arXiv preprint arXiv:2012.13349}.

\bibitem[{Parker and Rardin(2014)}]{parker2014discrete}
Parker, R.~G.; and Rardin, R.~L. 2014.
\newblock \emph{Discrete optimization}.
\newblock Elsevier.

\bibitem[{Poojari and Beasley(2009)}]{HeuristicMaster}
Poojari, C.; and Beasley, J. 2009.
\newblock Improving benders decomposition using a genetic algorithm.
\newblock \emph{European Journal of Operational Research}, 199(1): 89--97.

\bibitem[{Rahmaniani et~al.(2017)Rahmaniani, Crainic, Gendreau, and Rei}]{rahmaniani2017benders}
Rahmaniani, R.; Crainic, T.~G.; Gendreau, M.; and Rei, W. 2017.
\newblock The Benders decomposition algorithm: A literature review.
\newblock \emph{European Journal of Operational Research}, 259(3): 801--817.

\bibitem[{Schulman et~al.(2016)Schulman, Moritz, Levine, Jordan, and Abbeel}]{Schulmanetal_ICLR2016}
Schulman, J.; Moritz, P.; Levine, S.; Jordan, M.; and Abbeel, P. 2016.
\newblock High-Dimensional Continuous Control Using Generalized Advantage Estimation.
\newblock In \emph{Proceedings of the International Conference on Learning Representations (ICLR)}.

\bibitem[{Schulman et~al.(2017)Schulman, Wolski, Dhariwal, Radford, and Klimov}]{PPOSchulman}
Schulman, J.; Wolski, F.; Dhariwal, P.; Radford, A.; and Klimov, O. 2017.
\newblock Proximal policy optimization algorithms.
\newblock \emph{arXiv preprint arXiv:1707.06347}.

\bibitem[{Sutton and Barto(2018)}]{sutton_reinforcement_2018}
Sutton, R.~S.; and Barto, A.~G. 2018.
\newblock \emph{Reinforcement Learning: An Introduction}.
\newblock The MIT Press, second edition.

\bibitem[{Sutton et~al.(1999)Sutton, McAllester, Singh, and Mansour}]{SuttonPolicy}
Sutton, R.~S.; McAllester, D.; Singh, S.; and Mansour, Y. 1999.
\newblock Policy Gradient Methods for Reinforcement Learning with Function Approximation.
\newblock In Solla, S.; Leen, T.; and M\"{u}ller, K., eds., \emph{Advances in Neural Information Processing Systems}, volume~12. MIT Press.

\bibitem[{Tang, Agrawal, and Faenza(2020)}]{tang2020reinforcement}
Tang, Y.; Agrawal, S.; and Faenza, Y. 2020.
\newblock Reinforcement learning for integer programming: Learning to cut.
\newblock In \emph{International conference on machine learning}, 9367--9376. PMLR.

\bibitem[{Wang et~al.(2023)Wang, Li, Wang, Kuang, Yuan, Zeng, Zhang, and Wu}]{wang2023learning}
Wang, Z.; Li, X.; Wang, J.; Kuang, Y.; Yuan, M.; Zeng, J.; Zhang, Y.; and Wu, F. 2023.
\newblock Learning Cut Selection for Mixed-Integer Linear Programming via Hierarchical Sequence Model.
\newblock In \emph{The Eleventh International Conference on Learning Representations}.

\bibitem[{Zou and Hastie(2005)}]{zou2005regularization}
Zou, H.; and Hastie, T. 2005.
\newblock Regularization and variable selection via the elastic net.
\newblock \emph{Journal of the Royal Statistical Society Series B: Statistical Methodology}, 67(2): 301--320.

\end{thebibliography}

\section{Appendix}

\subsection{Inventory Management Problem}
\subsubsection{SO Formulation and Decomposition}\label{computational_evaluation}

To model the IMP as a SO mixed-integer problem we introduce the following notation: let $T$ be set of days $t$, $R$ be set of scenarios $r$, and  $S$ define a finite set of schedules $s$. Holding cost of an item (per unit-of-measure, per day) is $h$, the cost of emergency services (per unit) is $e$, the penalty applied to over-stocking (per unit over-stocked) is $q$, and $f_s$ is the fixed cost of a schedule. Capacity is defined by $m$ and starting inventory by $y$. The parameter $w_{st}$ indicates whether schedule $s$ orders on day $t$. Demand on day $t$ under scenario $r$ is $n_{tr}$.

The decision space is defined by seven sets of variables. The decision to use schedule $s$ is made using variable $u_s \in \{0,1\}$. The order-up-to amount is decided by  $a_{t} \in \mathbb{Z}^+$, and $k_{tr} \in \mathbb{Z}$ is the required order quantity to meet the order-up-to amount. Inventory on hand is monitored by  $d_{tr} \in \mathbb{Z}^+$, the units of holding space required to stock the inventory is $p_{tr} \in \mathbb{Z}^+$, the required emergency order quantity is $o_{tr} \in \mathbb{Z}^+$, and $v_{tr} \in \mathbb{Z}^+$ is the number of units that inventory is over-filled by (all defined $\forall t \in T, \forall r \in R$). The formulation of our IMP is

\begin{equation}\label{eq: mipobjective}
    min \sum_{\forall s \in S}(u_s f_s) + \frac{1}{R} \sum_{\forall r \in R} \sum_{\forall t \in T}(p_{tr}h + o_{tr}e + v_{tr}q)
\end{equation}

subject to:
\begin{equation}\label{eq: cashbal0}
    d_{tr} = y - n_{tr} + k_{tr} - v_{tr} + o_{tr},  t=0, \forall r \in R
\end{equation}
\begin{equation}\label{eq: cashbal}
    d_{tr} = d_{t-1, r} + k_{tr} - n_{tr} + o_{tr} - v_{tr}, \forall t \in \{1,...,T\}, \forall r \in R
\end{equation}
\begin{equation}\label{eq: hcstart}
    p_{tr} \geq y + k_{tr} - v_{tr}, t=0, \forall r \in R
\end{equation}
\begin{equation}\label{eq: hc}
    p_{tr} \geq a_t, \forall t \in \{1,...,T\}, \forall r \in R
\end{equation}
\begin{equation}\label{eq: hcdel}
    p_{tr} \geq p_{t-1, r} - a_t, \forall t \in \{1,...,T\}, \forall r \in R
\end{equation}
\begin{equation}\label{eq: cap0}
    y + k_{tr} - v_{tr} \leq m,  t=0, \forall r \in R
\end{equation}
\begin{equation}\label{eq: cap}
    d_{t-1,r} + k_{tr} - v_{tr} \leq m, \forall t \in \{1,...,T\}, \forall r \in R
\end{equation}
\begin{equation}\label{eq: order0}
    k_{tr} = a_{t} - y\sum_{\forall s \in S}u_sw_{st}
\end{equation}
\begin{equation}\label{eq: orderlb}
    k_{tr} \geq a_{t} - d_{t-1,r}, \forall t \in \{1,...,T\}, \forall r \in R
\end{equation}
\begin{multline}\label{eq: orderub}
    k_{tr} \leq a_{t} - d_{t-1,r} + (1-\sum_{\forall s \in S}u_sw_{st})m, \\ \forall t \in \{1,...,T\}, \forall r \in R
\end{multline}
\begin{equation}\label{eq: ordernosch}
    k_{tr} \leq a_{t}, \forall t \in T, \forall r \in R
\end{equation}
\begin{equation}\label{eq: negordernosch}
    k_{tr} \geq -\sum_{\forall s \in S} u_sw_{st} \times m, \forall t \in T, \forall r \in R
\end{equation}
\begin{equation} \label{eq: overfillonorder}
    v_{tr} \leq a_t, \forall t \in T, \forall r \in R
\end{equation}
\begin{equation}\label{eq: outnosch}
    a_{t} \leq \sum_{\forall s \in S} u_sw_{st} \times m, \forall t \in T
\end{equation}
\begin{equation}\label{eq: onesch}
    \sum_{\forall s \in S} u_s = 1
\end{equation}

The objective \eqref{eq: mipobjective} minimizes the sum of planned schedule costs and the average of holding costs, emergency order costs, and over-fill costs across the $R$ scenarios. Flow constraints \eqref{eq: cashbal0} and \eqref{eq: cashbal} balance inflow and outflow of inventory through demand and deliveries. The holding cost is enforced by constraints \eqref{eq: hcstart}, \eqref{eq: hc}, and \eqref{eq: hcdel}. Constraints \eqref{eq: cap0} and \eqref{eq: cap} mandate that inventory cannot be filled beyond its capacity. Lastly, constraints \eqref{eq: order0}, \eqref{eq: orderlb}, \eqref{eq: orderub}, \eqref{eq: ordernosch}, \eqref{eq: negordernosch}, \eqref{eq: overfillonorder}, and \eqref{eq: outnosch} ensure an order exactly fills the inventory to the optimal order-up-to-amount, and that orders are only placed on scheduled days. \eqref{eq: onesch} guarantees exactly one schedule is selected.

For BD, we note that $\mathbf{a}$ and $\mathbf{u}$ are schedule and order-up-to decisions that must be made the same across all scenarios. As a result, $\mathbf{a}$, $\mathbf{u}$, \eqref{eq: outnosch}, and \eqref{eq: onesch} are contained in the MIMP while the remaining decision variables and constraints are delegated to the scenario specific SPs. We define the dual variables in line with their related constraints: $\boldsymbol{\alpha} \in \mathbb{R}$ [\eqref{eq: cashbal0}, \eqref{eq: cashbal}], $\boldsymbol{\gamma} \in \mathbb{R}^+$ [\eqref{eq: hcstart}, \eqref{eq: hc}], $\boldsymbol{\omega} \in \mathbb{R}^+$ \eqref{eq: hcdel}, $\boldsymbol{\phi} \in \mathbb{R}^+$ [\eqref{eq: cap0},\eqref{eq: cap}], $\boldsymbol{\xi^0} \in \mathbb{R}$ \eqref{eq: order0}, $\boldsymbol{\xi^{lb}} \in \mathbb{R}^+$ \eqref{eq: orderlb}, $\boldsymbol{\xi^{ub} \in \mathbb{R}^-}$ \eqref{eq: orderub}, $\boldsymbol{\sigma} \in \mathbb{R}^-$ \eqref{eq: ordernosch},  $\boldsymbol{\pi} \in \mathbb{R}^+$ \eqref{eq: negordernosch}, $\boldsymbol{\phi} \in \mathbb{R}^-$ \eqref{eq: overfillonorder}.\\

The resulting decomposition consists of a master problem
\begin{equation}\label{eq: mpobjective}
    \min_{a,u,\theta}\sum_{\forall s \in S}(u_s \times f_s) + \frac{1}{R} \sum_{\forall r \in R}\theta_r
\end{equation}

s.t.
\begin{equation}\label{eq: outnoschmp}
    a_{t} \leq \sum_{\forall s \in S} u_sw_{st} \times m, \forall t \in T
\end{equation}
\begin{equation}\label{eq: oneschmp}
    \sum_{\forall s \in S} u_s = 1
\end{equation}
\begin{equation}\label{eq: zerotheta}
    \theta_r \geq 0, \forall r \in R    
\end{equation}

and dual sub-problem (solved independently for each scenario $r$):
\begin{multline}\label{eq: spobjective}
    \max_{\alpha, \phi, \xi^0, \xi^{lb}, \xi^{ub}, \sigma, \pi}  \alpha_{0r}(y - n_0r) + \gamma_{0r}y + \phi_{0r} (m - y) +\\
    \xi^0_r (a_0 - y \times \sum_{\forall s \in S} u_s w_{s0}) + \sum_{t=1}^T(-\alpha_{tr} n_{tr} + \gamma_{tr}a_t - \omega_{tr}a_t + \\
    \phi_{tr} m + 
    \xi^{lb}_{tr} a_{tr} +
    \xi^{ub}_{tr} (a_{tr} + (1 - \sum_{\forall s \in S} u_sw_{st})m)) + \\
    \sum_{\forall t \in T} (\phi_{tr} a_t + \sigma_{tr} a_t - \pi_{tr}(\sum_{\forall s \in S}u_sw_{st} \times m))
\end{multline}
s.t.
\begin{equation}
    \alpha_{tr} \leq 0, t=T
\end{equation}
\begin{equation}
    \alpha_{tr} - \alpha_{t+1, r} + \phi_{t+1, r} + \xi^{ub}_{t+1, r} + \xi^{lb}_{t+1, r} \leq 0, \forall t \in \{0,...,T-1\}
\end{equation}
\begin{equation}
    \gamma_{tr} - \omega_{t+1, r} \leq h, t = 0
\end{equation}
\begin{equation}
    \gamma_{tr} + \omega_{tr} \leq h, t = T
\end{equation}
\begin{equation}
    \gamma_{tr} - \omega_{t+1, r} + \omega_{tr} \leq h, \forall t \in \{1,...,T-1\}
\end{equation}
\begin{equation}
    \alpha_{tr} + \phi_{tr}  - \phi_{tr} + \gamma_{tr} \leq q, t = 0
\end{equation}
\begin{equation}
    \alpha_{tr} + \phi_{tr}  - \phi_{tr} \leq q, \forall t \in \{1,...,T\}
\end{equation}
\begin{equation}
    - \alpha_{tr} \leq p, \forall t \in T
\end{equation}
\begin{equation}
    \xi^0_r - \gamma_{tr} + \phi_{tr} - \alpha_{tr} + \sigma_{tr} + \pi_{tr} = 0, t = 0
\end{equation}
\begin{equation}
    \xi^{lb}_{tr} + \xi^{ub}_{tr} + \phi_{tr} - \alpha_{tr} + \sigma_{tr} + \pi_{tr} = 0, \forall t \in \{1,...,T\}
\end{equation}

\subsubsection{RL Formulation}
We define the following MDP for the IMP problem. The state is given by the tuple $s_t = \langle d, h, e, q, m, \mu, \sigma, \mathbf{w}, \mathbf{o}, \mathbf{r} \rangle \in \mathcal{S}$, where $t$ is a time step over the horizon $T$. Parameters $d$, $h$, $e$, $q$, $\mathbf{w}$, and $m$ directly follow the definitions introduced in the SO Formulation and Decomposition section of this Appendix. Additional state parameters include $\mu$ as the expected demand, and $\sigma$ as the estimated variance of demand. A vector $\mathbf{o}$ tracks orders over the time horizon $T$. All future orders are set to zero, and past orders are taken from actions as they are performed. Similarly, a vector $\mathbf{r}$ tracks the forecast errors from past observations. All future error observations are set to zero, and events are populated as the MDP steps through time.

The actions are given by $a_t = \langle \mathbf{k}_t, \mathbf{u}_t \rangle \in \mathcal{A}$ which denote the quantity to order and the schedule to adhere to at time $t$, respectively. Note that the schedule must be selected at the beginning of the horizon, and thus only $\mathbf{u}_{t=0}$ is relevant. This is enforced through action masking and for simplicity we will refer to $\mathbf{u}_{t=0}$ as $\mathbf{u}$. The reward is the negative cost, i.e. the negative of the objective defined in \eqref{eq: mipobjective}. 

As previously mentioned, we use PPO to train our actor-critic agent. We use a single multi-layer perception with action and value heads, with four hidden layers. The linear output layers for the actor return the log-odds, and the critic returns the value. We standardize the observation vector using a moving average during training, and generate $\mathbf{k}_t$ and $\mathbf{u}_t$ sequentially $\forall t \in T$.

The agent is presented with an initial state $s_0$ and must select a scheduling action to take. The scheduling action, $\mathbf{u}$, relates to a binary vector $\mathbf{w} \in \{0, 1\}^{T}$ that defines whether an order is possible on day $t$. If $\mathbf{w}_t = 1$, an order can be placed, otherwise the agent cannot order. For $t>0$ the selected schedule becomes part of the state, over-writing the initial zero vector $\mathbf{w}$.

With the schedule selected, the agent must generate a second action for state $s_0$; this time selecting an order amount. The repeated visitation of state $s_0$ is necessary as the selected schedule $\mathbf{w}$ has now become part of the state. While we have not temporally shifted the day, the state has changed.

After a second visitation of $s_0$, the agent sequentially traverses the horizon $T$. With each time step, an ordering decision is made and either accepted or masked depending on the schedule vector $\mathbf{w}$. Updates to the state include population of order quantities, residual updates, and an update of the inventory on hand based on observed demand and order amounts. If an order is scheduled, we retrieve the order-up-to amounts, denoted as $\mathbf{a}$ in the SO Formulation section, by adding inventory on hand at the end of $t-1$ to the ordering decision $k_t$. If an order is not scheduled the order-up-to amount is 0. After traversing the full horizon $T$, the agent will have selected a schedule $\mathbf{u} \in \{0, 1\}^S$ and have a vector of order-up-to quantities $\mathbf{a} \in \mathbb{Z}_{\geq 0}^T$ from the agent. These two decision vectors, $\mathbf{u}$ and $\mathbf{a}$ are the decisions required by the BD sub-problem. With these vectors, we can solve \eqref{eq: spobjective}, generate sub-gradient approximations on $\theta_r$, and further refine our approximation of the (stochastic) SP cost.

\subsection{Regularized Regression}
\subsubsection{Data Generation}
We assume data of the form $y = X^T \beta + \epsilon$ where $X$ is the feature set, $\beta$ is an unbiased vector of coefficients, and $\epsilon$ is Gaussian noise. To perform our experiments, we set the total number of features $P=10$ and generate $X \in \mathbb{R}^{(250 \times 10)}$ as a set of i.i.d. observations $X_{rj} \sim N(0, 1)$. 

We draw a value $k$ from a set of integer values $\mathbb{Z} \in [3, 8]$ to define the number of \emph{important} features for which $|\beta_i| > 0$. We next randomly draw, without replacement, $k$ values from $\mathbb{Z} \in [1, 10]$ to define columns where $\beta_i > 0$, and subsequently draw a random value for $\beta_i \sim U(-10, 10)$ for those columns. For all columns that were not included in the $k$ draws, we set $\beta = 0$.

To this point, we have sampled $X$, the important feature subset, and the coefficient values for those features. To finalize the data generation, we compute $X^T\beta = y - \epsilon$. To simulate $\epsilon$ we draw $\epsilon \sim U(0.05\hat{\mu}_y, 0.25\hat{\mu}_y)$ where $\hat{\mu}_y$ is the sample mean of $y$. We add $\epsilon$ to $y$ to recover the dataset $y = X^T\beta + \epsilon$, where $\sum_{\forall i \in I} ||\beta_i||_0 = k$.

\subsubsection{RL Formulation}
We model the RR problem as a MDP, with state $s_t = \langle \beta, p_{\beta}, \beta\vert_z, z \rangle \in \mathcal{S}$ where $\beta$ corresponds to the regression coefficients obtained if all $P$ features are used for regression, $p_{\beta}$ are their corresponding p-values, $\beta\vert_z$ are the regression coefficients of the selected features given by $z$, and $z$ is the multi-hot encoding vector for the selected features. Thus, $\langle \beta, p_{\beta} \rangle$ remain constant during an episode, whereas $\langle \beta\vert_z, z \rangle$ change at every step. 

We use PPO to train our actor-critic agent for 1e7 steps, which consists of a multi-layer perceptron with common hidden layers [256, 256], an action head  with hidden layers [256, 128, 64], and value heads with hidden layers [256, 128, 64]. Action outputs consist of a categorical distribution with log-odds for all available features, with action masking enabled for the previously selected features during an episode as the MDP transitions. 

\subsection{Results of $\boldsymbol{L_0}$ and $\boldsymbol{L_1}$ Regularization}
A comparison between $L_0$ and $L_1$ regularization for the RR problem is shown in Table \ref{tab: L0_vs_L1_RR}, using the following metric definitions (for $M$ observations, $N$ features):
\begin{itemize}
    \item $\beta$ Recovery: $\frac{1}{K}\sum_{\forall i \in N} ||\beta_i|_0 - |\hat{\beta_i}|_0 |_1$, where $K$ is the true number of relevant features.
    \item $\beta$ MSE: $\frac{1}{N}\sum_{\forall i \in N} (\beta_i - \hat{\beta_i})^2$
    \item Prediction MSE: $\frac{1}{M}\sum_{\forall i \in M} (y_i - \hat{y}_i)^2$
\end{itemize}
\begin{table}[h]
\centering
\begin{tabular}{@{}llll@{}}
\toprule
                       &
                       $L_0, \lambda = 0.1$ & 
                       $L_1, \lambda = 0.1$ &
                       $L_1, \lambda = 0.5$\\ \midrule
$\beta$ Recovery   & \textbf{0.056}  & 0.317 & 0.058 \\
$\beta$ MSE & \textbf{0.009} & 0.014 & 0.135\\
Pred MSE & 2.692 & \textbf{2.644} & 3.955 \\
\bottomrule
\end{tabular}
\caption{Comparative performance of $L_0$ and $L_1$ regularization. All metrics are averaged over 250 problems.}
\label{tab: L0_vs_L1_RR}
\end{table}
%\vspace{-1mm}

\clearpage

\end{document}